\documentclass[10pt,twocolumn,letterpaper]{article}

\usepackage{cvpr}              % To produce the CAMERA-READY version
\usepackage[dvipsnames]{xcolor}

\definecolor{cvprblue}{rgb}{0.21,0.49,0.74}
\usepackage[pagebackref,breaklinks,colorlinks,citecolor=cvprblue]{hyperref}

\usepackage{cuted}
\usepackage{float}
\usepackage{microtype}

\def\paperID{4594} % *** Enter the Paper ID here
\def\confName{CVPR}
\def\confYear{2024}

\title{\vspace{-1.4em}NeRFiller: Completing Scenes via Generative 3D Inpainting\vspace{-0.5em}}
\author{Ethan Weber$^{1,2}$\; Aleksander Hołyński$^{1,2}$\; Varun Jampani$^1$\; Saurabh Saxena$^1$ \\ Noah Snavely$^1$\; Abhishek Kar$^1$\;Angjoo Kanazawa$^2$ \\
$^1$Google Research \hspace{1cm} $^2$UC Berkeley
}

\begin{document}
%\maketitle
\twocolumn[{%
    \renewcommand\twocolumn[1][]{#1}%
    \maketitle
    \centering
    \vspace{-2em}
    \newcommand{\teaserwidth}{1.0\textwidth}
    \includegraphics[width=\textwidth]{assets/figures/teaser_small.pdf}
    % \rule{\teaserwidth}{22em} 
\captionof{figure}{\textbf{NeRFiller.} We propose a generative 3D inpainting approach for scene or object completion. Given a 3D capture with incomplete regions (left), our approach completes scenes such as the incomplete teddy bear scan (top left) and deletes unwanted occluders such as the pillow and the price tag (bottom left). We can also control the completions using a reference inpainted exemplar (right) to guide the process.
\label{fig:teaser}
}%
    \vspace{1em}
}]

% \begin{strip}
% \centering
% \captionsetup{type=figure}
% \includegraphics[width=\textwidth]{assets/figures/teaser.pdf}
% \end{strip}

\iftrue
% \iffalse
\newcommand{\ethan}[1]{\textcolor{blue}{[EW: #1]}}
\else
\newcommand{\ethan}[1]{{\noindent}}
\fi

\newcommand{\Lnerf}{\mathcal{L}_\mathit{nerf}}
\newcommand{\VSPACEAMOUNT}{{\vspace{-1.5em}}}

\newcommand{\SECVSPACEAMOUNT}{{\vspace{-1em}}}
\begin{abstract}
% \vspace{-2em}
% \looseness=-1
We propose NeRFiller, an approach that completes missing portions of a 3D capture via generative 3D inpainting using off-the-shelf 2D visual generative models. Often parts of a captured 3D scene or object are missing due to mesh reconstruction failures or a lack of observations (e.g., contact regions, such as the bottom of objects, or hard-to-reach areas). We approach this challenging 3D inpainting problem by leveraging a 2D inpainting diffusion model. We identify a surprising behavior of these models, where they generate more 3D consistent inpaints when images form a 2$\times$2 grid, and show how to generalize this behavior to more than four images. We then present an iterative framework to distill these inpainted regions into a single consistent 3D scene. In contrast to related works, we focus on completing scenes rather than deleting foreground objects, and our approach does not require tight 2D object masks or text. We compare our approach to relevant baselines adapted to our setting on a variety of scenes, where NeRFiller creates the most 3D consistent and plausible scene completions. Our project page is at \href{https://ethanweber.me/nerfiller}{https://ethanweber.me/nerfiller}.
\end{abstract}

%\vspace{-13mm}
\section{Introduction}
\label{sec:intro}
Consider the 3D scanned teddy bear and cat in Figure~\ref{fig:teaser}. In many 3D captures such as these, parts of the scene may not be as one desires: there may be unobserved regions such as the bottom of the bear and behind the cat, or there may be unwanted parts such as the price tag on the cat ear. Additionally, one may want to modify a feature, or generate a variety of alternative models, \eg, a bear with bunny ears or a santa cat. All of these tasks require the ability to edit and inpaint content in a 3D-aware and multi-view consistent manner. This is a challenge, since 2D generative inpainting models will not by default generate 3D consistent images. Our goal is to take a step in this direction and present a method that can create new content via scene completion conditioned on a set of multi-view images.

\begin{figure}[t]
\centering
\includegraphics[width=\linewidth]{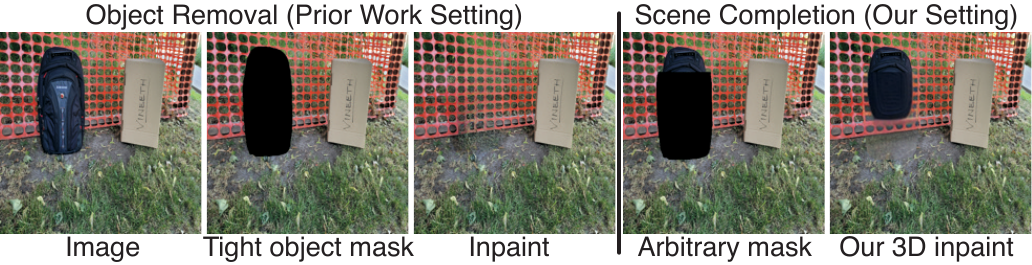}
\caption{\textbf{Object removal vs.\ scene completion.} We focus on scene completion (right) as opposed to object removal (left). Prior work focuses on removing entire objects with tight masks, while we tackle the more general setting of completing scenes with arbitrary missing regions across wide baselines. More realistic scenarios include missing regions or parts of scenes to edit, as illustrated in Figure~\ref{fig:teaser}.}
\label{fig:problem_setting}
\VSPACEAMOUNT
\end{figure}

Specifically, we present a 3D scene completion framework called \textit{NeRFiller}, which given a scene and specified parts of the scene to inpaint, returns a 3D scene that is completed in a multi-view consistent manner. Our approach not only completes missing regions (Figure~\ref{fig:teaser}, center), but can also generate multiple variations of the missing regions (Figure~\ref{fig:teaser}, right). Furthermore, our approach does not require text prompting and can operate from the scene context alone.

\begin{figure}[t]
\centering
\includegraphics[width=\linewidth]{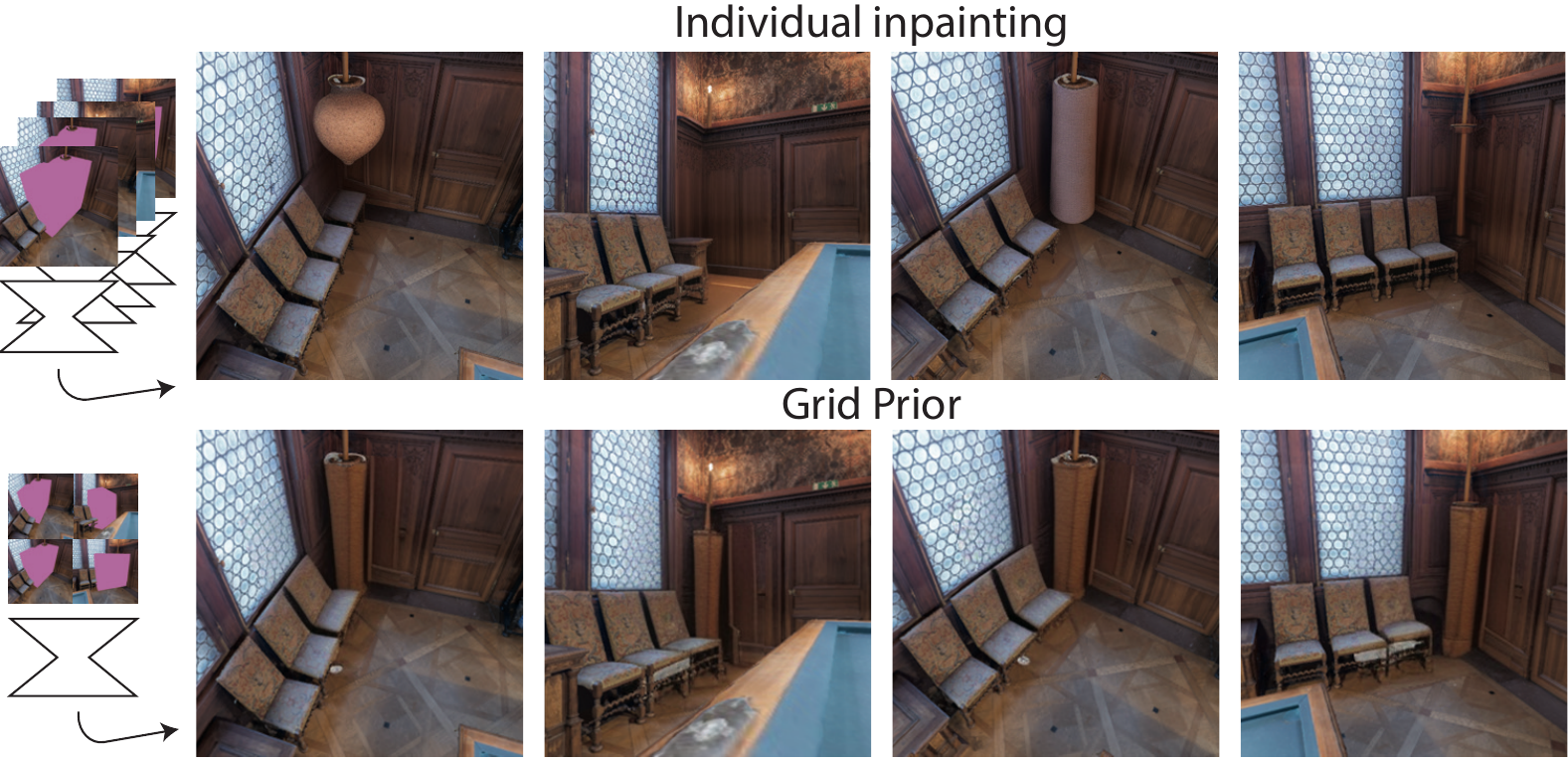}
\caption{\textbf{Grid Prior.} Here we inpaint the corner of the room (left illustrated in pink) with individual inpainting (top) and our Grid Prior method (bottom). Individual inpaints are diverse, while the Grid Prior encourages multi-view consistency.}
\label{fig:grid_prior}
\VSPACEAMOUNT
\end{figure}

We achieve this by proposing a novel approach to generate inpaints with an off-the-shelf 2D generative image model in a manner that encourages multi-view consistency. Specifically, we identify a useful phenomenon in text-to-image diffusion models that we refer to as a \textit{Grid Prior}: denoising four images with missing observations that are tiled in a 2$\times$2 grid results in more consistent multi-view inpaints than inpainting them independently, shown in Figure~\ref{fig:grid_prior}. We propose a method called \textit{Joint Multi-View Inpainting} that generalizes this behavior to more than four images. While this technique results in more 3D consistent inpaints, it is still a 2D-based approach and 3D consistency is not guaranteed. Therefore, we propose a way to distill these inpaints in a global 3D scene representation in an iterative manner.

While there has been a surge of recent works that generate 3D scenes completely from scratch using text~\cite{fridman2023scenescape,hollein2023text2room} or image guidance~\cite{liu2021infinite,li2022infinitenature}, our approach differs in that we focus on completing scenes given the context of an existing 3D scene. Our approach is related to recent methods that remove a specified object from a scene~\cite{spinnerf}, but we can generate new content that goes beyond completing a textured background, as illustrated in Figure~\ref{fig:problem_setting}. We also do not assume a tight object mask, and can generate a diverse set of inpaints.

To demonstrate the efficacy of our approach, we experiment with a diverse set of scenes including 3D indoor photogrammetry captures lacking coverage in certain areas, 3D scenes with specified missing regions, and 3D objects. While our problem is challenging, we show that NeRFiller can recover more 3D consistent and plausible results compared to recent state-of-the-art methods adapted to our setting.

\section{Related Work}
Our goal is to complete missing parts of an existing 3D scene. There are several ways to approach this, via 2D inpainting or via distilling a 2D generative model for 3D generation.

\medskip\noindent\textbf{2D inpainting.} 2D inpainting methods take an image and mask and complete the missing content at the mask location. Early methods relied on inpainting by copying texture from known regions into the unknown regions~\cite{efros1999texture}. A state-of-the-art model is LaMa (Large Mask inpainting)~\cite{suvorov2021resolution__lama}, which is particularly good at infilling large missing areas. It uses fast Fourier convolutions, a large receptive field, and large training masks. This model is highly effective at completing plausible ``background'' textures within a specified mask (Fig.~\ref{fig:problem_setting} left) but lacks diverse outputs as it is deterministic. Probabilistic diffusion models~\cite{ho2020denoising, po2023state} have recently produced remarkable results for image generation. They can also be used for inpainting and can generate diverse inpainted outputs. Pixel-based diffusion models do not have to be trained explicitly for inpainting, but can be modified at test-time by setting known regions before each denoising iteration~\cite{lugmayr2022repaint}. Latent diffusion models (LDMs)~\cite{rombach2022high__sd_stable_diffusion} are also effective at inpainting and are efficient because they operate in latent space. However, they require fine-tuning for inpainting with image and mask conditioning. 2D inpainting models can be prompted~\cite{bar2022visual} and/or fine-tuned~\cite{tang2023realfill} enabling additional flexibility for downstream applications.

\begin{figure*}[t]
\centering
\includegraphics[width=\linewidth]{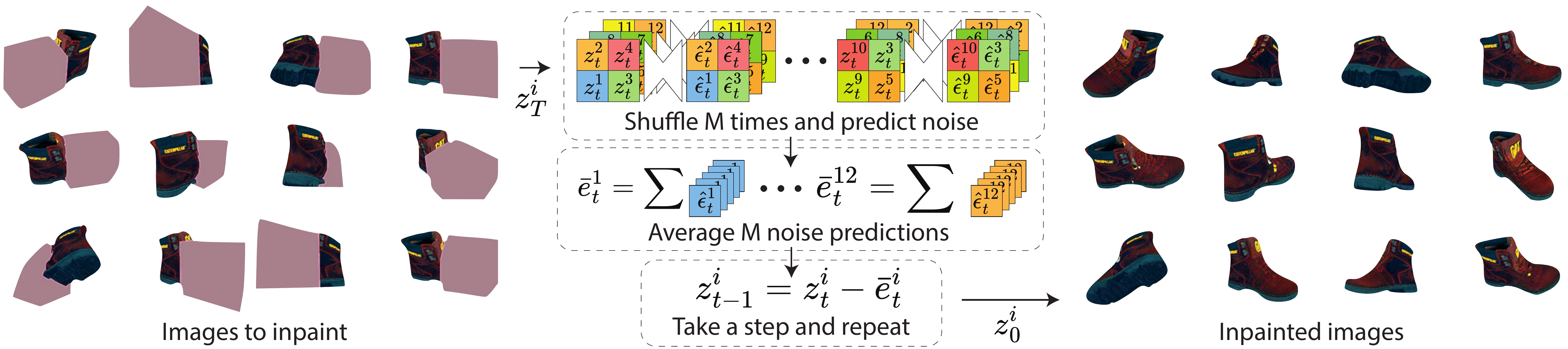}
\caption{\textbf{Joint Multi-View Inpainting.} We enable properties of the Grid Prior with more than four images by averaging diffusion model predictions. We take $N$ images (left), create $N/4$ grids, and obtain a noise prediction from SD~\cite{rombach2022high__sd_stable_diffusion}. We do this $M$ times and average the noise predictions before taking a denoising step. At $z_{0}$, the images (right) are fairly consistent and can be used to train a NeRF with our Inpaint DU method.}
\label{fig:joint_inpainting}
\VSPACEAMOUNT
\end{figure*}

\medskip\noindent\textbf{3D generation.} 3D generation takes as input text or images and outputs 3D content. The Infinite Nature line of work~\cite{liu2021infinite,li2022infinitenature,koh2021pathdreamer,wiles2020synsin,rockwell2021pixelsynth}, takes as input a single image and generates immersive fly-through content using a 2D inpainting model queried in an autoregressive manner~\cite{liu2021infinite,li2022infinitenature}. Cai et. al.~\cite{cai2022diffdreamer} follow this path with a diffusion model, however, none of these approaches can recover a global 3D scene representation. Persistent Nature~\cite{chai2023persistent} and related work~\cite{chen2023scenedreamer,devries2021unconstrained} maintain a latent scene but are completely generative and not conditioned on input image sets. SceneScape~\cite{fridman2023scenescape} and Text2Room~\cite{hollein2023text2room} use text prompts and 2D inpainters to create a 3D mesh by using an inpainter and depth predictor to successively stitch a mesh. These approaches cannot fix a mistake in the scene if a bad inpaint is made during the successive stitching because no global optimization is performed.

Other methods create 3D content via a global optimization strategy. DreamFusion~\cite{poole2022dreamfusion} and related works~\cite{wang2023prolificdreamer,zhou2023sparsefusion} use the NeRF framework to optimize a 3D volume given a text prompt. Others train models to have 3D consistent properties \cite{zhou2023sparsefusion,watson2022novel,tang2023mvdiffusion,li2023panogen,liu2023syncdreamer}. Follow-up works leverage 2D diffusion models techniques~\cite{ruiz2023dreambooth,gal2022image__textualinversion} to create 3D content conditioned on real images~\cite{raj2023dreambooth3d,melas2023realfusion,chan2023generative}. These approaches are not designed for the inpainting task.

\medskip\noindent\textbf{3D inpainting.} Unlike most 3D generation methods, we ground our inpaints with an actual 3D scene or object that has missing regions (Figure~\ref{fig:teaser}). Casual capture~\cite{mildenhall2019llff,broxton2020immersive,single_view_mpi,Casual3D2017,Instant3D2018} or NeRFs~\cite{mildenhall2021nerf,kerbl3Dgaussians} is a use-case as they often contain artifacts when rendered from novel views~\cite{Nerfbusters2023,goli2023}. Most relevant to scene the completion setting is the object removal setting (see Figure~\ref{fig:problem_setting}). These works remove foreground objects from NeRF captures~\cite{wang2023inpaintnerf360,spinnerf,Weder2023Removing}. They do this by inpainting each image in a NeRF dataset once and training with various losses including patch-based perceptual losses and depth regularization. \cite{mirzaei2023reference} enables inpainting from a reference image. A variety of these methods are evaluated on the SPIn-NeRF dataset, which is in the forward-facing LLFF~\cite{mildenhall2019llff} format and has small parallax. Our work uses datasets with a significantly larger baseline.

Our focus is on the more general scene completion setting, which is related to editing. IN2N~\cite{instructnerf2023} edits a scene using InstructPix2Pix~\cite{brooks2022instructpix2pix}, but it cannot hallucinate new geometry. The video editing literature is also relevant, with techniques such as extended attention from Tune-A-Video~\cite{wu2023tune,tokenflow2023} to encourage consistency in edited video frames. However, editing 2D images does not guarantee consistency when lifted to 3D. In our method, we encourage 2D inpaints to converge via iterative NeRF optimization and dataset updates. Moreover, we achieve this with \textit{off-the-shelf} 2D generative models without the need for expensive purpose-trained diffusion models or model fine-tuning.

\section{Preliminaries}

\subsection{Neural Radiance Fields (NeRFs)}
Neural radiance fields (NeRFs)~\cite{mildenhall2021nerf} represent the 3D geometry and radiance of a scene with neural networks. NeRFs take as input an 3D position ($x, y, z$) and a viewing direction ($\theta, \phi$), and output a color and density ($c, \sigma)$. To train a NeRF $f_{\Theta}$, a set of calibrated, posed images are used to construct a set of 3D rays $\mathbf{r}(t) = \mathbf{o} + t \mathbf{d}$ for each pixel with known color $C(\mathbf{r})$. During training, these rays are sampled and rendered via volumetric rendering to obtain a color estimate $\hat C(\mathbf{r})$. Rays are sampled from training images and the field is optimized with photometric losses $\Lnerf(C(\mathbf{r}), \hat C(\mathbf{r}))$, e.g., MSE or LPIPS~\cite{zhang2018perceptual__lpips}. During inference, a full image is rendered with all rays of the desired camera.

\begin{figure*}[t]
\centering
\includegraphics[width=\linewidth]{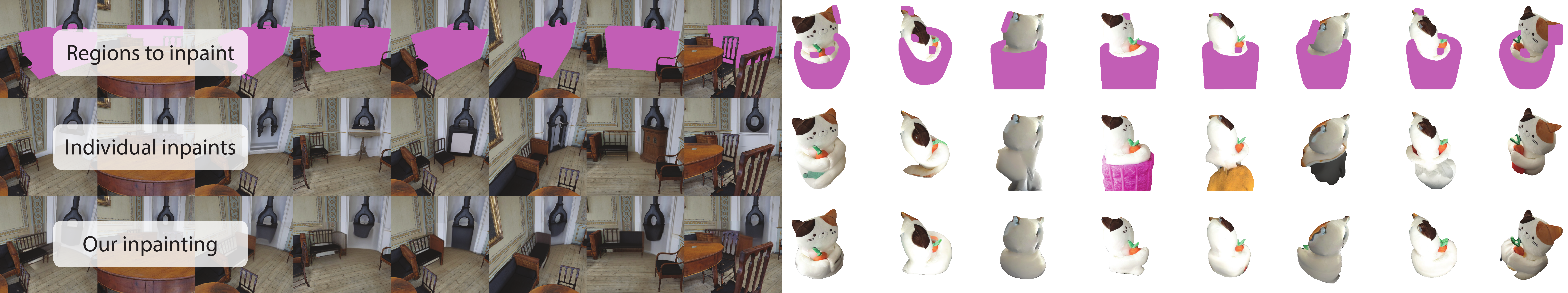}
\caption{\textbf{Joint Multi-View Inpainting Examples.} The top images are inpainted with SD~\cite{rombach2022high__sd_stable_diffusion} without any text conditioning (middle) and with our Joint Multi-View Inpainting method (bottom). Our joint inpaints are more multi-view consistent.}
\label{fig:joint_inpainting_examples}
\VSPACEAMOUNT
\end{figure*}

\subsection{2D Diffusion Models}
\label{sec:2d_diffusion_models}

Diffusion models consist of two processes: a forward process $q$ that gradually adds noise to a data sample $z_{0} \sim p_\mathit{data}(z)$, and a learned reverse process to iteratively denoise a pure Gaussian noise sample $z_{T} \sim \mathcal{N}(0, 1)$ into a clean image $z_{0}$. An intermediate noisy $z_{t}$ can be obtained from the clean image by adding noise $\epsilon$ with scaling $\bar \alpha_{t}$, where $z_{t} = \sqrt{\bar \alpha_{t}} z_{0} + \sqrt{1 - \bar \alpha_{t} \epsilon}$.
The diffusion model $\epsilon_\theta$ predicts noise $\hat{\epsilon}$ present in the image $z_{t}$ as $\hat \epsilon = \epsilon_{\phi} (z_t, t, c)$. $t$ is a time indicating how much noise is in the sample, and $c$ is a general form of conditioning (e.g., images, masks, or text). During training, random noise $\epsilon$ and $t$ are sampled and the objective $\mathcal{L}_\mathit{diff} = || \hat{\epsilon} - \epsilon ||^2$ is minimized. With the prediction $\hat \epsilon$ at time $t$, a reduced noise $z_{t-1}$ can be obtained by $z_{t-1} = z_{t} - \hat \epsilon$ (where we omit the scaling of $\hat \epsilon$ for simplicity). Repeating this until $z_{0}$ yields a fully denoised sample. Stochastically training with $c$ (conditionally) and without $c$ (unconditionally) enables classifier-free guidance (CFG)~\cite{ho2022classifier} during inference time.
% We leverage this property when sampling the 2D diffusion inpainting model.
In practice, $z$ is latents since we are using SD (Stable Diffusion)~\cite{rombach2022high__sd_stable_diffusion}, but in general we can map from $z$ to higher resolution pixels $x$ with an encoder $\mathcal{E}(x)$ and decoder $\mathcal{D}(z)$.

\medskip\noindent\textbf{Using a diffusion model as a prior.} Diffusion models have advantages over other models (e.g., GANs and deterministic inpainters~\cite{suvorov2021resolution__lama}) because they can be used a a prior to optimize underlying variables such as the parameters of a 3D NeRF $f_\Theta$
% ~\cite{poole2022dreamfusion,sjc} 
with methods like score distillation sampling (SDS)~\cite{poole2022dreamfusion,sjc}. When used as a prior for NeRFs, the objective is to find the best $\Theta$ such that a rendered image $x$ has high likelihood under the diffusion model prediction $\epsilon_{\phi}$. SDS involves rendering an image,
% encoding it (for latent diffusion models),
adding partial noise, and updating the NeRF such that the diffusion model can predict the added noise. 
IN2N~\cite{instructnerf2023} introduced a variant of this method coined Dataset Update (DU). Instead of backpropping based on the diffusion model prediction, DU renders an image, adds partial noise, and takes multiple steps to recover an estimated clean image $x_{0}$.
The clean image is added to the dataset and used to supervise the NeRF. Every $S$ iteration, another image is replaced. The DU supervision signal will be slightly delayed since images are cached for several iterations, while SDS provides immediate gradients corresponding to the current render. However, we use the DU method in our work because it has a few advantages over SDS in terms of implementation: We can obtain higher-resolution supervision (albeit slightly delayed) with less GPU memory and we can update a large batch of images simultaneously (e.g., 40 images). We find that large batch updates are important for our inpainting task since we are changing the NeRF geometry, unlike prior works that focus purely on modifying appearance~\cite{instructnerf2023,nguyen2022snerf}.

\section{Method}
Our method, NeRFiller, aims to complete a missing region within a 3D scene by using an inpainting prior from a generative 2D diffusion model. This problem statement poses a number of challenges. 
First, the inpainted estimates from a 2D diffusion model are diverse, and may vary from sample to sample. This requires a consolidation mechanism to ensure that the completed 3D scene contains one salient inpainted result, as opposed to the average of all possibilities. 
Second, 2D inpainting models are not trained for 3D consistency, and will therefore provide estimates that cannot be explained by a single 3D scene, even if they correspond to the same approximate style or content. In the following, we describe our approaches to tackle these problems. In Section~\ref{sec:multi_view_inpainting}, we describe how we encourage the inpainted outputs from a diffusion sampling process to be 3D consistent. In Section~\ref{sec:inpaint_idu}, we describe an iterative 3D scene optimization method that uses these inpainted images to optimize for a globally consistent inpainted 3D scene. NeRFiller builds on an observation that inpainting a grid of images encourages the outputs to have similar appearance, and we extend this idea to an arbitrarily large collection of images through a joint sampling approach.

\begin{figure}[t]
\centering
\includegraphics[width=\linewidth]{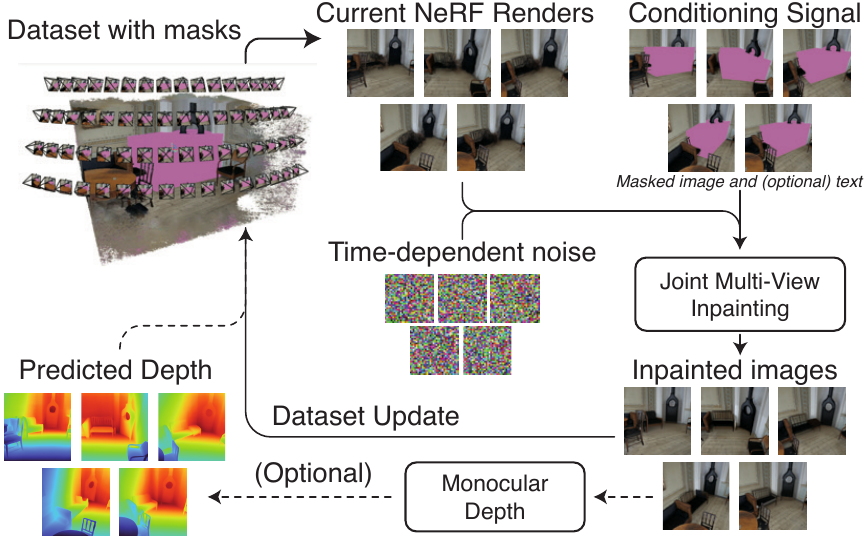}
\caption{\textbf{Inpaint Dataset Update.} Every $S$ iterations, we update the unknown pixels of the NeRF training images. We render $N$ images, add partial noise, and jointly inpaint with a conditioning signal. We (optionally) predict the depth and update the dataset.}
\label{fig:inpaint_dataset_update}
\VSPACEAMOUNT
\end{figure}

\begin{figure*}[t]
\centering
\includegraphics[width=\linewidth]{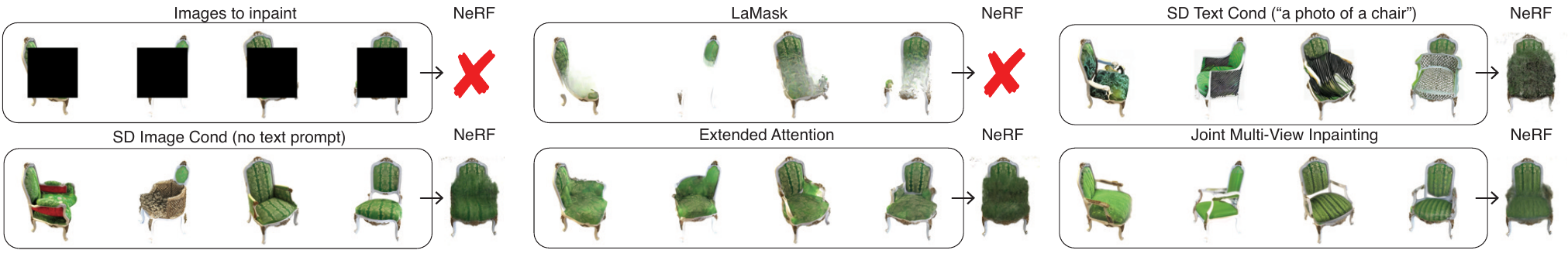}
\caption{\textbf{Inpainting methods.} Inpainting methods produce inconsistent inpaints. We show various inpainting methods (boxed) and use a collection of them to train a NeRF. A resulting render is shown on the right of each method. Using our Grid Prior and Joint Multi-View Inpainting creates reasonably consistent inpaints and a plausible NeRF. $\mathcal{X}$ means the NeRF failed and resulted in white everywhere.}
\label{fig:inpaint_methods}
\VSPACEAMOUNT
\end{figure*}

\subsection{Multi-view consistent inpainting}
\label{sec:multi_view_inpainting}
A core challenge in 3D inpainting with a 2D generative model is getting the outputs of the 2D model to be consistent across views. This challenge stems from the multimodality of the output distribution: in most cases, there are many plausible inpaintings, and sampling multiple 
% similar or otherwise 
consistent images remains an open research problem. 

\medskip\noindent\textbf{Grid Prior.} While inpainting multiple viewpoints independently may produce inconsistent results, one interesting discovery is that consistency can be achieved by tiling the input images into a grid and treating the \emph{grid} of images (and their corresponding masks) as a single inpainting target. This grid-based prior can produce more 3D consistent views, both in coarse appearance and approximate scene structure (illustrated in Figure~\ref{fig:grid_prior}). We hypothesize that this phenomenon results from similarly structured examples in Stable Diffusion's training dataset: sets of observations depicting the same scene or object organized as a grid (e.g., screenshots of online product photos). Similar properties were also explored in visual-prompting~\cite{bar2022visual}.

More specifically, in order to inpaint four images consistently, one can downsample them and their corresponding inpainting masks to quarter-resolution and tile them as a 2$\times$2 grid. This grid is fed through the 2D inpainting model (as a single image would) to get as output four inpainted images with consistent content. More formally, let $\mathcal{G}$ be the downsampling and grid operation for four images and let $\mathcal{G}^{-1}$ undo this. We can grid four images latents $z^{1}_{t}, z^{2}_{t}, z^{3}_{t}, z^{4}_{t}$ as follows:
%%%%%%
%% \vspace{-1em}
\begin{gather}
\{\hat \epsilon^{1}_t, \hat \epsilon^{2}_t, \hat \epsilon^{3}_t, \hat \epsilon^{4}_t\} = \mathcal{G}^{-1}(
\epsilon_{\phi}
(
\mathcal{G}(\{z^{1}_{t}, z^{2}_{t}, z^{3}_{t}, z^{4}_{t}\})
)
)
\end{gather}
%%%%%
and take a denoising step with $z^{i}_{t-1} = z^{i}_t - \hat \epsilon^{i}_t$. In principle, this approach is similar to recent methods that use \emph{extended attention}, i.e., shared keys and values in the attention operations across a set of parallel sampling processes~\cite{wu2023tune}. Our approach does not share attention features but instead shares context with other images via the diffusion U-Net receptive field that sees 4 tiled images at a time. In our experiments, we compare to \emph{extended attention} and demonstrate that our grid prior more effectively inpaints 3D consistent content when used with our Joint Multi-View Inpainting method.

% \scriptsize
\begin{table}[t]
    \captionsetup{font=small}
    \centering
    \scriptsize
    \begin{tabular}{l|lll} % the number of columns needed
        \toprule
        
 & PSNR $\uparrow$ & SSIM $\uparrow$ & LPIPS $\downarrow$ \\
        \midrule
        
 Masked NeRF & 7.76 & 0.71 & 0.37 \\ 
 LaMask & 19.58 & 0.89 & 0.20 \\ 
 SD Text Cond & 12.56 & 0.73 & 0.32 \\ 
 SD Image Cond & 14.15 & 0.76 & 0.28 \\ 
 Extended Attention & 14.57 & 0.77 & 0.27 \\ 
 Grid Prior & 14.43 & 0.80 & 0.25 \\ 
 Joint Multi-View Inpainting & 15.89 & 0.82 & 0.23 \\ 

        \bottomrule
    \end{tabular}
    \caption{\textbf{Multi-view consistent inpainting.} 
We inpaint images and train a NeRF for the 8 scenes of the NeRF synthetic dataset. Better metrics indicate more consistency of the NeRF 3D reconstruction with the 2D inpaints. Note that LaMask achieves the best results as it often copies the white background into the hole (see Figure~\ref{fig:inpaint_methods}) and results in a failed NeRF that is totally white.
}
    \label{tab:2d_inpainting_consistency}
\end{table}

\medskip\noindent\textbf{Joint Multi-View Inpainting.} While effective at inpainting a set of images consistently, applying the grid prior to a larger set of images poses additional challenges. Increasing the number of images in the grid proportionally decreases the output resolution of each image (e.g., arranging a set of 2$\times$2 images in a grid reduces each image's resolution by 4, a 3$\times$3 grid by 9, and so on). For the purpose of producing high-quality inpainting results, we would like to minimize any loss in image detail. Therefore, we propose a method that uses the above grid prior in a joint sampling process, inspired by MultiDiffusion~\cite{bar2023multidiffusion}. In each sampling step, we shuffle all the input images into a set of 2$\times$2 grids. We repeat this $M$ times and before taking a sampling step, the score estimate for each image is combined across all the grid combinations in which it was seen. This causes the inpainting estimates to be gradually shared across the entire dataset, effectively increasing the grid size without further reducing effective resolution.

\begin{figure*}[t]
\centering
\includegraphics[width=\linewidth]{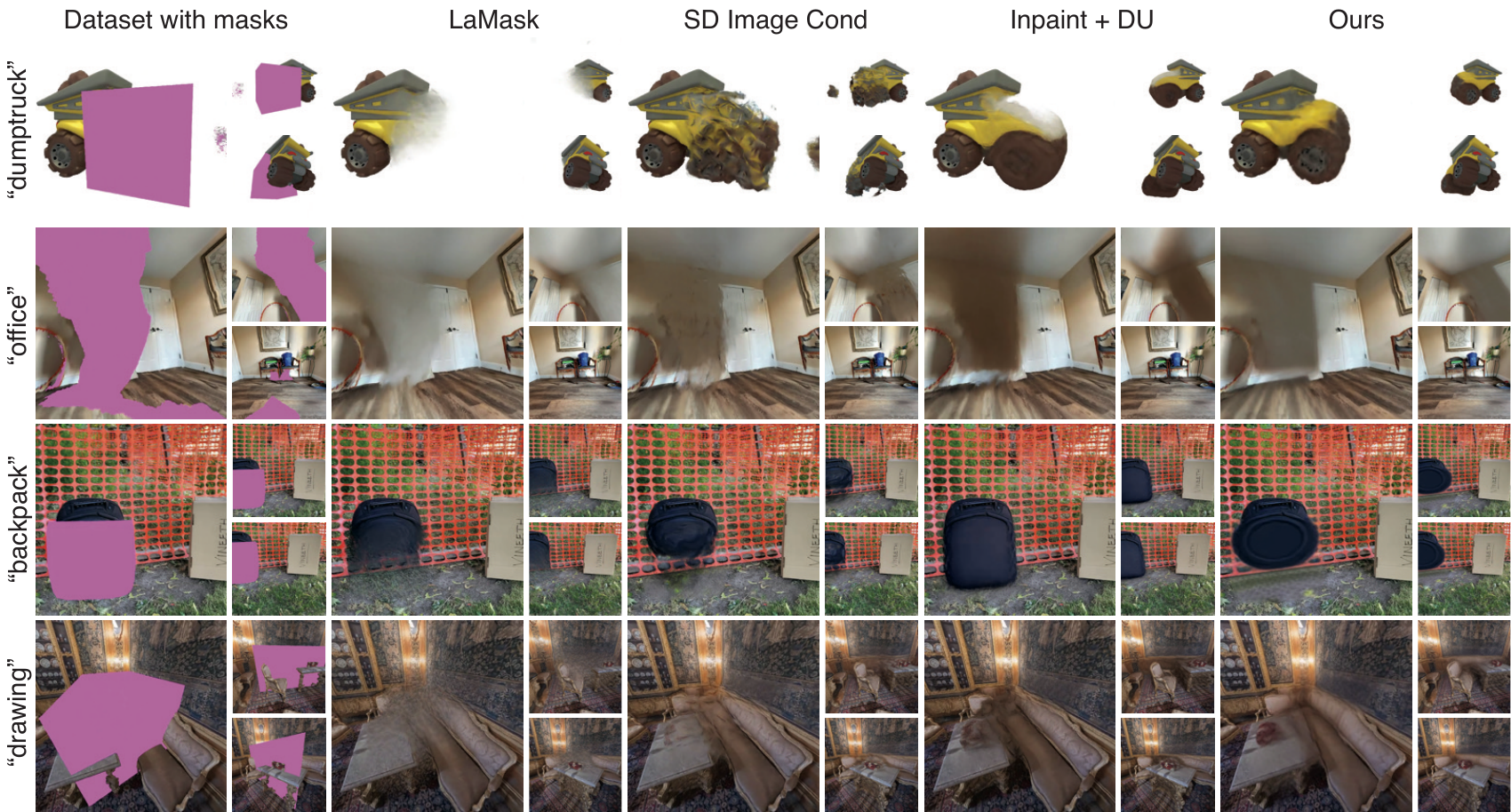}
\caption{\textbf{Qualitative NeRF results.} On the left, we show various scenes with pink regions to be completed. We compare NeRFiller (far right) against baselines adapted to our scene completion setting. The ``office" scene is missing parts of the wall, floor, and under the chairs.}
\label{fig:qualitative_nerf_baselines}
\VSPACEAMOUNT
\end{figure*}

Figure~\ref{fig:joint_inpainting} describes this procedure. More formally, for a batch of $N$ images, we randomly permute their order, construct $N/4$ grids, and predict the noise for $M$ iterations ($j \in [1, M]$), as follows:
%%%%%
%% \vspace{-1em}
\begin{gather}
\{\hat \epsilon^{1j}_t \ldots\ \hat \epsilon^{Nj}_t\} = \mathcal{G}^{-1}(
\epsilon_{\phi}
(
\mathcal{G}(\{z^{1j}_{t} \ldots\ z^{Nj}_{t}\})
)
)
\end{gather}
%%%%%%
and step with $z^{i}_{t-1} = z^{i}_t - \sum_{j \in M} \hat \epsilon^{ij}_t$ from $z_{T}$ to $z_{0}$. Qualitative results are shown in Figure~\ref{fig:joint_inpainting_examples}.

\subsection{Completing 3D Scenes}
\label{sec:inpaint_idu}
The proposed Joint Multi-View Inpainting enables inpainting images in a more 3D consistent manner than other inpainting methods. Next, we describe how to distill these 2D inpainting results into a single 3D reconstruction. 
We refer to our method as Inpaint Iterative Dataset Update (Fig.~\ref{fig:inpaint_dataset_update}), or \textbf{Inpaint DU}, as it derives from IN2N's~\cite{instructnerf2023} Iterative DU method (see Sec.~\ref{sec:2d_diffusion_models}). In contrast to IN2N, which begins with a complete NeRF reconstruction and uses a model conditioned on complete 2D observations, our task requires us to train a complete NeRF from images with masked unknown regions. As in IN2N, we begin training with a dataset of original (known) pixels, and update the dataset over training by adding or replacing the set of initially unknown pixels with the inpainted estimates. Fig.~\ref{fig:inpaint_dataset_update} illustrates this procedure. Specifically, every $S$ steps, we render the set of $N$ training views, encode them into latents $z^{i}_{0}$ and then partially noise them before feeding them to SD. We sample from these partially noised inputs using the proposed Joint Multi-View Inpainting strategy, then use the resulting images to replace the corresponding images in the dataset. This process is both prefixed and suffixed by an encode and decode operation, since the base model is a latent diffusion model. We repeat this process many times while linearly annealing $t$ from full noise $t=1$ to $t=t_\mathrm{min}$. In practice, we set $t_\mathrm{min} = 0.4$, since we find that low noise values result in quality degradation. We observe that over the course of optimization, our inpainted images become gradually more consistent (Fig.~\ref{fig:inpaints_over_time}) as the added geometry and texture begin to take form. Annealing $t$ helps encourage the inpainting to converge to a single result rather than making large changes late in training.

\medskip\noindent\textbf{Depth regularization.} Inpaint DU optionally incorporates depth supervision to improve inpainted scene geometry. After each dataset update, we predict the depth for all images with ZoeDepth~\cite{bhat2023zoedepth}. We use a relative depth ranking loss~\cite{wang2023sparsenerf} in the inpainted regions (but not on the known pixels). We use a ranking loss because it's a softer constraint than metric depth supervision, where errors in scale-and-shift alignment could more easily harm the 3D scene geometry. \textit{We only apply depth supervision for our main method to indoor scenes and not objects, since we empirically noticed that \cite{bhat2023zoedepth} performs less consistently when the background is a solid color (e.g., white or black).}

% \scriptsize
\begin{table}[t]
    \captionsetup{font=small}
    \centering
    \scriptsize
    \begin{tabular}{l|lll|ll} % the number of columns needed
        \toprule
        
 & PSNR $\uparrow$ & SSIM $\uparrow$ & LPIPS $\downarrow$ & MUSIQ $\uparrow$ & Corrs $\uparrow$ \\
        \midrule
        
 Masked NeRF & 14.71 & 0.78 & 0.26 & 3.71 & 675 \\ 
 LaMask & 27.39 & 0.90 & 0.05 & 3.76 & 643 \\ 
 SD Image Cond & 22.03 & 0.86 & 0.11 & 3.68 & 665 \\ 
 Inpaint + DU & 26.60 & 0.89 & 0.08 & 3.76 & 660 \\ 
 Ours w/o depth & 28.41 & 0.92 & 0.06 & 3.72 & 682 \\ 
 Ours & 28.28 & 0.91 & 0.06 & 3.73 & 696 \\ 

        \bottomrule
    \end{tabular}
    \caption{\textbf{Quantitative NeRF results.} 
We report various metrics averaged over our 10 scenes to quantify consistency of the 3D NeRF with the dataset inpaints (left), as well as novel-view metrics (right), such as ``Corrs" (number of high-quality correspondences between random pairs of frames) for geometry.
}
    \label{tab:quantitative_nerf_baselines}
\end{table}

\section{Experiments}
We compare NeRFiller for 3D scene completion to various inpainting baselines. We first investigate various inpainting strategies on multi-view synthetic scenes to gauge how effective a deterministic inpainter~\cite{suvorov2021resolution__lama} is compared to SD~\cite{rombach2022high__sd_stable_diffusion}, sampled in various ways. After establishing that our \textit{Joint Multi-View Inpainting} demonstrates multi-view inpainting properties, we evaluate our full method on 10 scans with missing regions. We compare NeRFiller to various object-removal baselines, \textit{adapted to our setting}, to complete missing regions. Finally, we analyze the parameters of our method and show an application to reference-guided scene completion. We conduct experiments with Nerfstudio~\cite{tancik2023nerfstudio} and provide implementation specifics in the appendix.

\subsection{3D consistent image inpainting}
\label{sec:3D_consistent_image_inpainting}

Our goal is to evaluate various 2D inpainting models and strategies to quantify their 3D consistency.

\medskip\noindent\textbf{Setting and evaluation.} For these experiments, we take the testing split of the NeRF synthetic dataset~\cite{mildenhall2021nerf} (8 scenes of 200 images each). We resize each image to 512$\times$512 resolution and mask out the center of each image with a 256$\times$256 region to inpaint (see Figure~\ref{fig:inpaint_methods} top left). We inpaint all images with various methods and train a Nerfacto NeRF model~\cite{tancik2023nerfstudio} on 180 equally spaced images and evaluate metrics on the remaining 20 images. We use the standard NeRF metrics because they capture how similar the 3D reconstruction is to the 20 hold-out evaluation images. When rendering for evaluation (right of inpainted images in Figure~\ref{fig:inpaint_methods}), we push the near plane slightly forward to avoid including any floaters hiding in front of the cameras.
% Please see the supmat for videos.

\medskip\noindent\textbf{Baselines.} Our baselines are the following:

\begin{itemize}
  \item \textit{Masked NeRF} - No inpainting, only train on known pixels.
  
  \item \textit{LaMask} - LaMa~\cite{suvorov2021resolution__lama} inpainting model.

  \item \textit{SD Text Cond} - SD using a text CFG with prompt ``a photo of \{description\}". See the appendix for text prompts.
  
  \item \textit{SD Image Cond} - SD with only image CFG.
  
  \item \textit{Extended Attention} - SD with only image CFG and extended attention~\cite{wu2023tune}.
\end{itemize}
%%%%%
\begin{figure}[t]
\centering
\includegraphics[width=\linewidth]{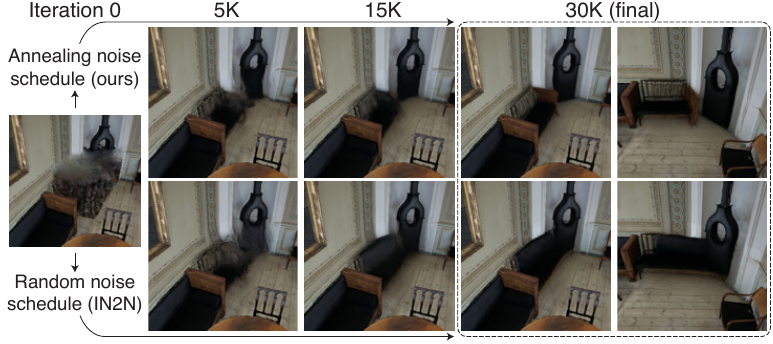}
\caption{\textbf{Noise schedule.} We anneal the amount of noise we add to the NeRF renders when making a dataset update, while I-N2N chooses random noise each time. Annealing the noise produces sharper results (top) while I-N2N's noise schedule introduces significant blur (bottom).}
\label{fig:inpaints_over_time}
\VSPACEAMOUNT
\end{figure}
%%%%%
For \textit{Extended Attention} and \textit{Grid Prior}, we inpaint in batches of 5 and 4, respectively, and for \textit{Joint Multi-View Inpainting}, we inpaint 40 images simultaneously with $M=8$ diffusion averaging steps. To inpaint additional batches of 40 images, we set 20 in the batch as known. This enables fitting within the memory constraints of a 16 GB GPU.

\medskip\noindent\textbf{Results.} Our results are shown in Tab.~\ref{tab:2d_inpainting_consistency}, where we see that \textit{Joint Multi-View Inpainting} achieves the best metrics in multi-view consistent inpainting. Our results show that we have achieved some level of multi-view consistency. Our images look the most consistent in Figure~\ref{fig:inpaint_methods} (bottom right), and the trained NeRF looks plausible. The other methods yield significant blur in the NeRF reconstruction. We provide videos on the project page showing the NeRF results for each method.

\subsection{Completing large unknown 3D regions}
\label{sec:completing_large_unknown}

In this setting, our goal is to complete missing regions in 3D content. We construct a set of 10 datasets consisting of various 3D content. For some scenes e.g., the backpack from~\cite{spinnerf}, we modify their provided mask to include part of the object (Fig.~\ref{fig:problem_setting}) to convert it to the scene completion setting. For other scenes, we simply want to fill in any missing details (e.g., parts of walls). Some of the meshes are missing vertices after multi-view stereo reconstruction, e.g., ``bear" and ``office". For others, we place a large 3D occluder in the scene to simulate the scene completion setting. The pink regions in Fig.~\ref{fig:qualitative_nerf_baselines} (left) shows the areas to complete. We create the datasets by rendering $\sim$60 novel views looking at the occluded region. Importantly, the rendered images have enough known pixels to provide context to the inpainting model that the images observe the same scene from different camera viewpoints. Our datasets have much more parallax than the forward-facing scenes of ~\cite{spinnerf}. Our appendix provides details on our data, including where we obtained our 3D content, mostly from Objaverse~\cite{deitke2023objaverse,deitke2023objaversexl} and Sketchfab.

\medskip\noindent\textbf{Evaluation.} The evaluation is similar to Sec.~\ref{sec:3D_consistent_image_inpainting} for the dataset images. However, in this case our task is to construct a scene for good novel-view synthesis, so we use all images for both training and evaluation. We compute NeRF metrics on the entire images, where we compare the final rendered images with the latest version of the inpainted region. For methods that inpaint once without DU (e.g., \textit{LaMask}), we compare against the first and only inpaints. For DU methods, we compare against the latest round of inpaints. \textit{Our metrics are against inpainted images which serves to evaluate the consistency of the scene because there is no ground-truth solution.} We also report novel-view metrics, computed on a custom 10 second 30 FPS camera path novel views that moves around the scene. We also report an image quality metric MUSIQ~\cite{ke2021musiq} and a geometry metric. For geometry, we report the number of high-quality LoFTR~\cite{sun2021loftr} correspondences between 100 randomly sampled pairs of frames. More high-quality matches should correlate with better multi-view consistency and fewer extreme view-dependent effects that destroy realism. Please see the Appendix for more details.

\begin{figure}[t]
\centering
\includegraphics[width=\linewidth]{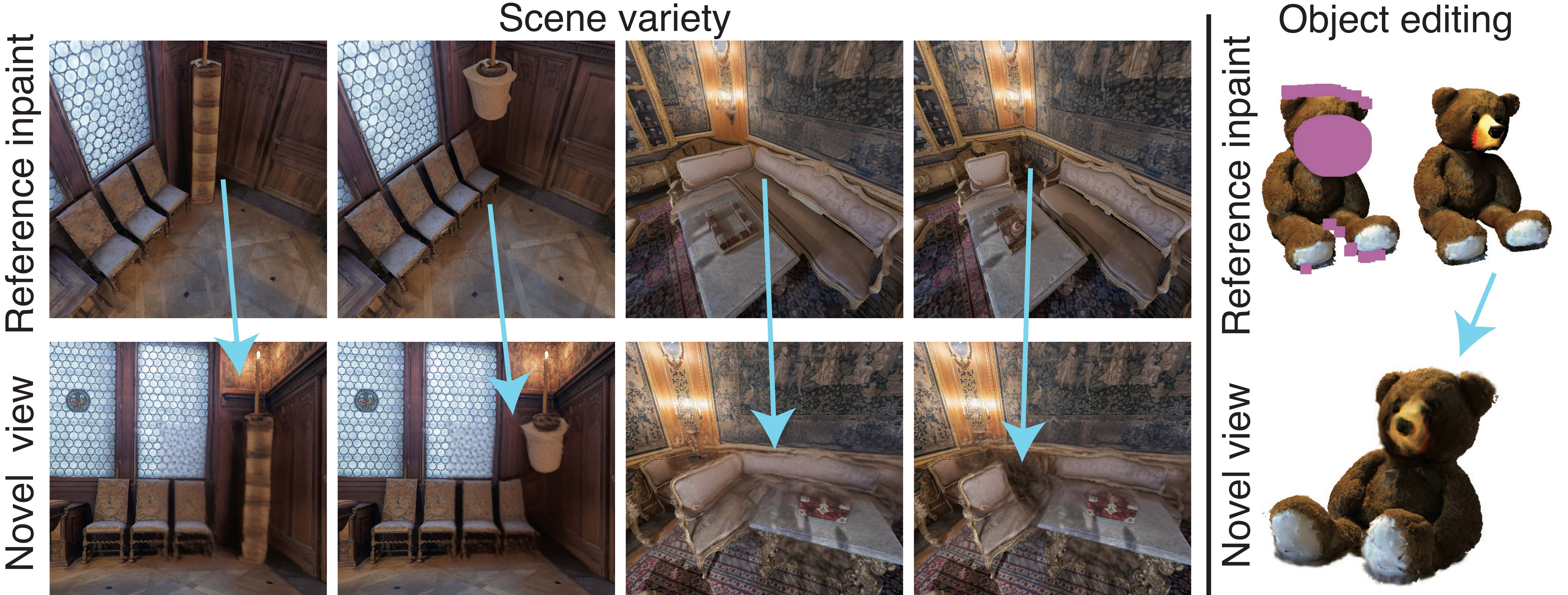}
\caption{\textbf{Reference-based completion.} Given a reference inpaint (top row), we propagate it into a 3D NeRF (bottom row).}
\label{fig:reference}
\VSPACEAMOUNT
\end{figure}

\medskip\noindent\textbf{Baselines.} We implemented the following baselines:

\begin{itemize}
  \item \textit{Masked NeRF} - no inpainting, where we train a Nerfacto model~\cite{tancik2023nerfstudio} only in the known pixel locations, 
  
  \item \textit{LaMask} - Inpaint once with LaMa~\cite{suvorov2021resolution__lama} and train with patch-based perceptual losses. This is our adaptation of SPIn-NeRF~\cite{spinnerf}.
  
  \item \textit{SD Image Cond} - Inpaint once with SD. This is similar to InpaintNeRF360~\cite{wang2023inpaintnerf360} but without text since we find in Sec.~\ref{sec:3D_consistent_image_inpainting} that text CFG produces very inconsistent inpaints.

  \item \textit{Inpaint + DU} - An adaption of IN2N~\cite{instructnerf2023} for our setting, which inpaints one image at a time with the SD inpainting model and our annealed noise schedule.

\end{itemize}

\medskip\noindent\textbf{Results.} Some qualitative results are shown in Fig.~\ref{fig:qualitative_nerf_baselines} and full videos for all 10 scenes are provided in the appendix. \textit{LaMask} and \textit{SD Image Cond} are both inpaint-once methods and therefore create large blurry regions in the NeRF, but between the two, \textit{LaMask} is smoother since its deterministic inpainter \cite{suvorov2021resolution__lama} is less creative than SD in its outputs. LaMa~\cite{suvorov2021resolution__lama} tends to copy background textures into the mask region. From certain views, \textit{Inpaint + DU} looks sharp due to inpainting individually at full resolution; however, it has geometric inconsistencies and view-dependent effects which are crisp from some angles and blurry in others. Our method looks the most consistent with plausible outputs, although its ability to create consistent high-frequency texture details may be improved. We provide quantitative results showing that our final renders are most similar with the latest round of inpaints (Tab.~\ref{tab:quantitative_nerf_baselines} left). For novel-view metrics, we obtain the most correspondences (Tab.~\ref{tab:quantitative_nerf_baselines} right). Note that although we make an effort to capture the results quantitatively, there is no singular ground truth and therefore the results are best discerned qualitatively by viewing videos.

\subsection{Reference-based inpainting}
\label{sec:reference_guided_inpainted}
In some situations it is desirable to have control over the content used to complete the scene. NeRFiller can be easily adapted to 3D inpaint with respect to a user-provided reference inpaint. To do this, we first inpaint from one view and use it to prompt our Grid Prior update method. We ensure that each grid has the one reference inpaint when passed through SD. This ensures that all U-Net predictions are influenced by the reference inpaint so that new inpaints are more likely to be consistent with the reference. For this, we edit 30 images at a time (instead of 40) and make 10 grids, each with exactly 1 reference inpaint. See Figure~\ref{fig:teaser} and Figure~\ref{fig:reference} for examples.

\begin{figure}[t]
\centering
\includegraphics[width=\linewidth]{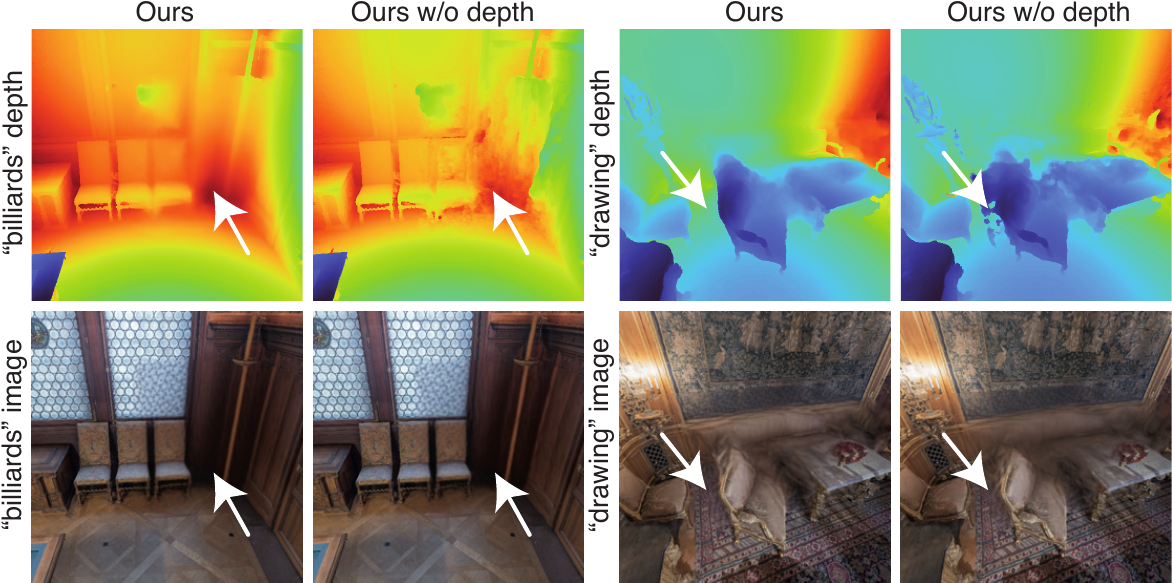}
\caption{\textbf{Relative depth supervision.} Relative depth supervision cleans up geometry (top) without affecting visual quality (bottom).}
\label{fig:depth_supervision}
\VSPACEAMOUNT
\end{figure}

\subsection{Parameter choices}
\vspace{-0.5em}
\medskip\noindent\textbf{Noise schedule.} It is important to anneal the amount of noise added the rendered images. We start by adding full noise (1.0) and decrease it to 0.4 over the 30K iterations of training. IN2N~\cite{instructnerf2023}, in contrast, uses a random schedule of choosing between 0.98 and 0.02 each update. This likely works because the InstructPix2Pix model is image conditioned and geometry of the NeRF does not change much. In our case, we are changing the geometry and using their schedule leads to blurry results, shown in Figure~\ref{fig:inpaints_over_time}.

\medskip\noindent\textbf{Depth regularization.} We find that adding depth ranking supervision~\cite{wang2023sparsenerf} in \textit{Ours} improves geometry but it hardly changes the quantitative or visual results. Our method without depth supervision is still favorable compared to the baselines. In Figure~\ref{fig:depth_supervision}, we see that the geometry is significantly improved but the RGB NeRF renderings are nearly indistinguishable. Consequently, we use depth supervision on indoor scenes since having better geometry is favorable for downstream applications such as mesh export.

\section{Limitations}
\label{sec:limitations}
%\medskip
\noindent\textbf{Low resolution and blur.} Our method recovers coarse geometry quite well but struggles to recover high-resolution detail in regions far way from the training cameras. We suspect this is because our \textit{Joint Multi-View Inpainting} method downsamples images to construct the $2\times2$ grids. Perhaps a post-processing method to fine-tune SD~\cite{sun2023dreamcraft3d} or a GANeRF~\cite{roessle2023ganerf} like optimization could improve the fidelity, but recovering high-frequency details remains a fundamental issue for 3D generative methods, e.g., DreamFusion~\cite{poole2022dreamfusion}.

%\medskip
\noindent\textbf{Inpainting NeRF casual captures}. An application of our approach would be to inpaint deleted content from Nerfbusters~\cite{Nerfbusters2023} or Bayes' Rays~\cite{goli2023}. However, these masked-out regions are large (limiting scene context to an inpainter) and furthermore, their mask patterns cause SD~\cite{rombach2022high__sd_stable_diffusion} to fail without multiple iterations of dilation, as pointed out in Text2Room~\cite{hollein2023text2room} and in our appendix. One could retrain SD with these mask distributions, but this is out of scope of our method which uses an \textit{off-the-shelf} model.

\section{Conclusion}
In this paper, we propose a generative 3D inpainting method called NeRFiller, which leverages an \textit{off-the-shelf} 2D inpainting model~\cite{rombach2022high__sd_stable_diffusion} to complete missing parts of 3D scenes and objects. We discover a unique property of these models where tiling four images into a 2$\times$2 grid produces more consistent inpaints than inpainting them independently. We exploit this property and propose \textit{Joint Multi-View Inpainting},  which enables inpainting many images simultaneously with more consistency by averaging noise predictions. %\textit{Joint Multi-View Inpainting} 
We show how to use it in the NeRF setting by performing iterative dataset updates. We evaluate against relevant state-of-the-art baselines adapted to our problem setting on a variety of 3D captures. Our approach also enables users to specify how to fill in the missing regions. Many 3D captures are incomplete with holes, and our work presents a framework for completing these missing regions.

\section*{Acknowledgements}

This project is supported in part by IARPA DOI/IBC 140D0423C0035. The views and conclusions contained herein are those of the authors and do not represent the official policies or endorsements of IARPA, DOI/IBC, of the U.S. Government. We would like to thank Frederik Warburg, David McAllister, Qianqian Wang, Matthew Tancik, Grace Luo, Dave Epstein, Riley Peterlinz for discussions and technical support. We also thank Ruilong Li, Evonne Ng, Adam Rashid, Alexander Kristoffersen, Rohan Mathur, Jonathan Zakharov for proofreading drafts and providing feedback.

{
    \small
    \bibliographystyle{ieeenat_fullname}
    \bibliography{main}
}

% \clearpage
\section*{Appendix}

The appendix includes more details related to the paper.

\section{NeRF synthetic prompts}

We use the following text prompts for each of the 8 NeRF synthetic datasets from the original NeRF paper~\cite{mildenhall2021nerf}.

\begin{itemize}
    \item ``chair": ``a photo of a chair"
    \item ``drums": ``a photo of drums"
    \item ``ficus": ``a photo of a ficus plant"
    \item ``hotdog": ``a photo of a hotdog"
    \item ``lego": ``a photo of a lego bulldozer"
    \item ``materials": ``a photo of materials"
    \item ``mic": ``a photo of a microphone"
    \item ``ship": ``a photo of a ship
\end{itemize}

\section{Datasets}

Some of our data can be found online with their respective hyperlinks. Others were created by the authors by scanning with Polycam, and one was obtained by modifying a SPIn-NeRF dataset.

\begin{itemize}
  \item \href{https://sketchfab.com/3d-models/dumptruck-qCWXrNLMlMOEtD5rcS0zOKbdkbB}{``dumptruck"} (Sketchfab)
  \item ``turtle" (author capture with Polycam)
  \item \href{https://sketchfab.com/3d-models/the-great-drawing-room-feb9ad17e042418c8e759b81e3b2e5d7}{``drawing"} (Sketchfab)
  \item \href{https://spinnerf3d.github.io/}{``backpack"} (SPIn-NeRF dataset)
  \item \href{https://sketchfab.com/3d-models/teddy-bear-459ce5b99da947438d3ea11d7c0d4225}{``bear"} (Sketchfab)
  \item \href{https://sketchfab.com/3d-models/the-billiards-room-79615d823a9149069dcd06c20bc9707f}{``billiards"} (Sketchfab)
  \item \href{https://sketchfab.com/3d-models/caspar-herman-storms-salong-ladegarden-2f0ad401c83041be9cc070316fdbc922}{``norway"} (Sketchfab)
  \item ``cat" (author capture with Polycam)
  \item \href{https://sketchfab.com/3d-models/caterpillar-work-boot-d551ce74dcd24528a05cbb0f4b7434d7}{``boot"} (Sketchfab)
  \item \href{https://sketchfab.com/3d-models/untitled-scan-288a4dec375d478a8286f4ef0ca42d06}{``office"} (Sketchfab)
\end{itemize}

\subsection{Marking inpaint regions}

To mark inpaint regions, we have a variety of approaches depending on the data. For the ``office" scene, we mark inpaint regions as any missing region from the original mesh, found \href{https://sketchfab.com/3d-models/untitled-scan-288a4dec375d478a8286f4ef0ca42d06}{here} on Sketchfab. For the other scenes, we place a 3D occluder, such as a cube or cylinders. For the ``cat" dataset in Figure~\ref{fig:teaser}, we use a cylinder to mark an occlusion of the body and a rectangular prism to mark the tag by the ear. It's quite straightforward to load the mesh into Blender and add the occluders. However, we leave the automatic placement of 3D occlusions for future work. Uncertainty masks from Bayes' Rays~\cite{goli2023} or deleted areas from Nerfbusters~\cite{Nerfbusters2023} could provide masks for our method, but their masks are out-of-distribution for SD (Stable Diffusion), shown in Fig.~\ref{fig:nerfbusters} and discussed in Sec.~\ref{sec:nerfbusters}.

\subsection{Creating NeRF datasets}

To create NeRF datasets with the marked inpaint regions, we render the mesh from $\sim$60 images (64 for all except ``backpack" which uses 60). For objects, we render around the object and for e.g., large missing rectangles from a room, we render arcs looking toward the region from different angles, at different elevations. We can mark the inpainting regions by doing a depth check between the actual mesh and the manually placed occluders. To create the ``backpack" scene, we took the dataset from SPIn-NeRF~\cite{spinnerf} and modified the original tight mask around the backpack to include the top part. We also dilated the masks to make them less tight. See Figure~\ref{fig:problem_setting} for an illustration of our changes to convert it for our scene completion setting rather than object deletion.

\section{Additional experiment details}

\subsection{Inpaint implementation details}

\medskip\noindent\textbf{Image classifier-free guidance.} Some percentage of the time during fine-tuning of the Stable Diffusion model for inpainting~\cite{rombach2022high__sd_stable_diffusion}, everything is masked out. This is similar to dropping out the text prompt with some probability to enable classifier-free guidance. Because of this training strategy, we can enable image-guidance with a formulation similar to InstructPix2Pix~\cite{brooks2022instructpix2pix}. We modify the text-conditioned diffusion inpainting model to predict its score estimate in the form:
%%%%
\begin{gather}
    \epsilon_{\phi}(z_t, t, c) = \epsilon_{\phi}(z_t, t, \{c_I, \emptyset\}) \nonumber\\
    + s_I \cdot (\epsilon_{\phi}(z_t, t, \{c_I, c_T\}) - \epsilon_{\phi}(z_t, t, \{c_I, \emptyset\})) \\
    + s_T \cdot (\epsilon_{\phi}(z_t, t, \{c_I, \emptyset\}) - \epsilon_{\phi}(z_t, t, \{\emptyset, \emptyset\})) \nonumber
\end{gather}
%%%%
with $z_t$ as the noisy latent at time $t$, $c_I$ and $c_T$ as conditioning for image \& mask, and text respectively. $\emptyset$ means no text or completely masking out the image. $s_I$ and $s_T$ are image and text guidance scales, respectively. We use $s_I > 0$ and $s_T = 0$ to make the model conditioned only on the image and masked region to inpaint. We note that the popular diffusers\footnote{\url{https://github.com/huggingface/diffusers}} library doesn't enable setting $s_I > 0$ \textit{out-of-the-box}, so most works use diffusion inpainting models by using text prompts to describe an inpaint. Trying to describe the scene, however, can be difficult and using image only guidance (``SD Image Cond", where $s_I > 0$ and $s_T = 0$) tends to be more multi-view consistent (Figure~\ref{fig:inpaint_methods}).

\medskip\noindent\textbf{Scheduler details.} We use DDIM for our scheduler. We use 20 steps whenever sampling SD~\cite{rombach2022high__sd_stable_diffusion}, regardless of how much noise is added to a current render. This is similar to Instruct NeRF2NeRF~\cite{instructnerf2023}'s dataset update procedure. The model we use can be found \href{https://huggingface.co/stabilityai/stable-diffusion-2-inpainting}{here} as ``stabilityai/stable-diffusion-2-inpainting" on Hugging Face.

\subsection{NeRF implementation details}

For synthetic scenes and objects, we adopt recommended practices for training Nerfacto~\cite{tancik2023nerfstudio} which is to set the background color to either white or black, disable scene contraction, and turn off the distortion loss. For the ``backpack" forward facing scene from SPIn-NeRF~\cite{spinnerf}, we only turn off the distortion loss. Nerfacto is not tuned for LLFF~\cite{mildenhall2019llff} forward-facing scenes, so there may be some artifacts in this dataset compared to our datasets which have larger parallax. For other scenes e.g., those that resemble an indoor room, we train Nerfacto with default settings.

For the baselines besides ``Masked NeRF" in Sec.~\ref{sec:completing_large_unknown} ``Completing large unknown 3D regions" and Table~\ref{tab:quantitative_nerf_baselines}, we train a modified Nerfacto that has losses on patches. Note that ``Masked NeRF" is simply Nerfacto, with any modifications as described. The other baselines render 1024 patches of size $32\times32$ per iteration. An L1 RGB loss and LPIPS~\cite{zhang2018perceptual__lpips} loss are applied to these patches, similar to IN2N~\cite{instructnerf2023}. Furthermore, we start the dataset update (DU) methods from the result of ``Masked NeRF". Each method is trained for 30K iterations, which is typical for Nerfacto.

\subsection{Evaluation against inpainted images}

We use $t \in [0.4, 1.0]$ noise added to a render before sampling SD and updating a training image. Because $t=0.4$ is the minimum noise, the inpainter has some freedom to change the inpaint from the current NeRF render. If we added no noise ($t=0.0$) to the current NeRF render, then the DU (dataset update) methods would be perfect on the NeRF metrics (PSNR, LPIPS, SSIM) because SD would do nothing and the inpainted images would be exactly the NeRF rendered images. Nevertheless, quantifying inpaints without a ground truth result is challenging and our metrics represent an effort to show quantitative evaluation of 3D consistency. We recommend looking at the videos to compare the methods qualitatively.

\subsection{Near plane for evaluation metrics}

We render with a near plane slightly in front of the training images when reporting our metrics. This is important because NeRFs can cheat by placing ``floaters" in front of the training images to explain away any discrepancies in the actual 3D scene compared to the training images. By setting the near plane a bit beyond the camera origin, we can render the true 3D scene without the floaters directly in front of training images.

\subsection{Novel-view video metrics}

For the \textit{MUSIQ} image quality metric, also used in~\cite{mirzaei2023reference}, we take all 300 frames (10 seconds, 30 FPS) of the novel-view videos, downsample each frame by 4x to capture low-frequency structure, and run the MUSIQ~\cite{ke2021musiq} model (trained on the AVA dataset) to obtain an image quality score. For the \textit{Corrs} metric, we use LoFTR~\cite{sun2021loftr} with a confidence threshold of $0.8$. We count the number of correspondences above this confidence threshold for 100 random pairs of frames.

\subsection{Metrics for each scene}

We provide the individual tables for Table~\ref{tab:2d_inpainting_consistency} and Table~\ref{tab:quantitative_nerf_baselines} of the paper. See the end of the document for these.

\section{Inpainting NeRF casual captures}
\label{sec:nerfbusters}

\begin{figure*}[p]
\centering
\includegraphics[width=\linewidth]{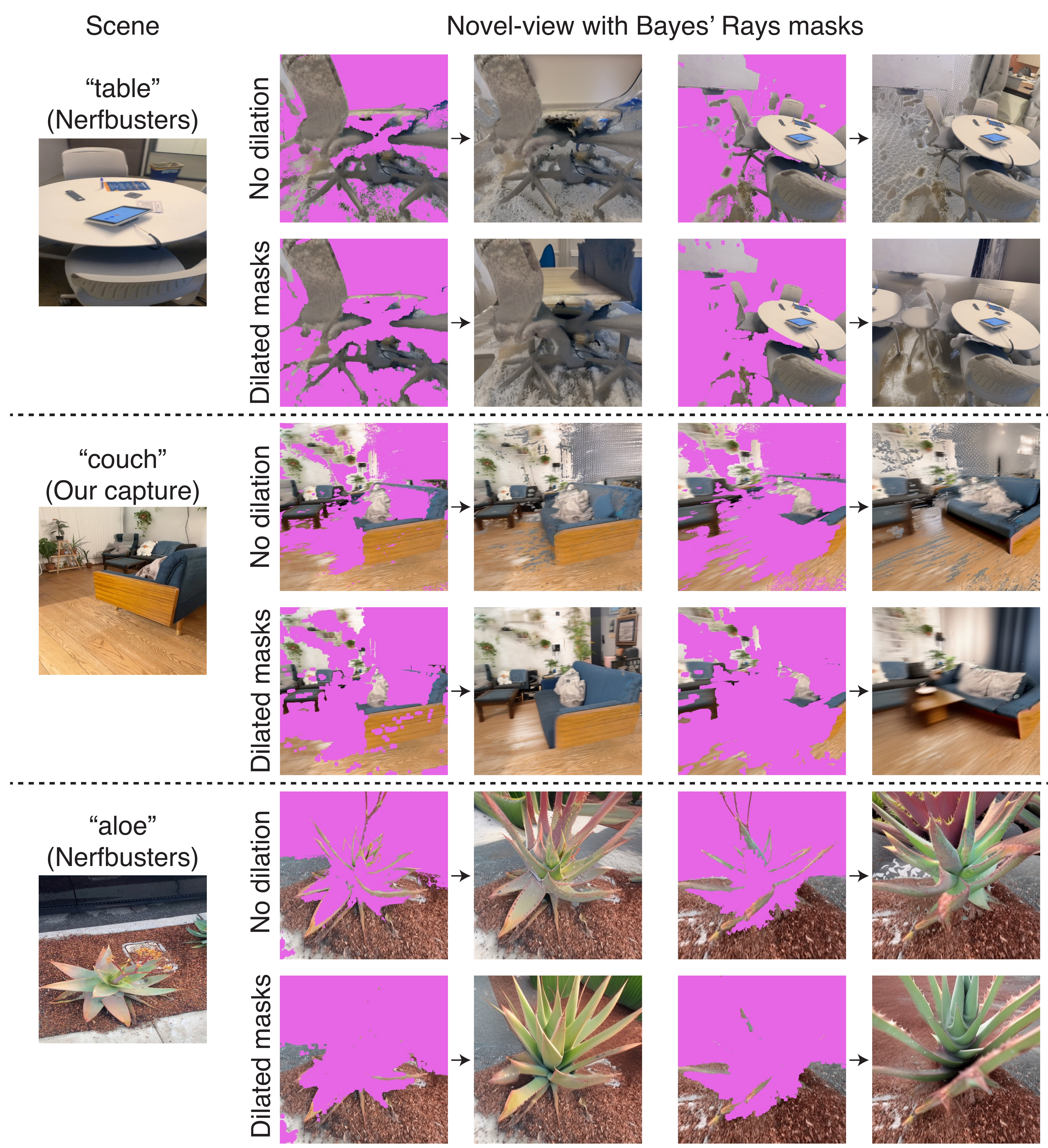}
\caption{\textbf{Out-of-distribution inpaints from Stable Diffusion when applied to casual captures.} On the left, we show a training image from a casually captured scene. On the right, we show masks obtained by rendering a novel-view that has occlusions, and then running Bayes' Rays~\cite{goli2023} to delete areas with high uncertainty (marked in pink). We then inpaint these regions with image conditioning. When the mask is not dilated (top rows), the inpaints have many artifacts such as ripple patterns and gray stretches. When the masks are dilated (bottom rows), the inpaints get slightly better but are no longer consistent with the known parts of the scene. This is a challenging setting to address in future work. Retraining SD with masks of this distribution could alleviate the problem, but this is costly and out-of-scope of using an \textit{off-the-shelf} model, as done in our work.}
\label{fig:nerfbusters}
\end{figure*}

As mentioned in our limitations section (Sec.~\ref{sec:limitations}), Stable Diffusion (SD) fails to produce good inpaints when the mask distribution is similar to those produced by Bayes' Rays~\cite{goli2023} or Nerfbusters~\cite{Nerfbusters2023}. In Fig.~\ref{fig:nerfbusters}, we illustrate this. Please see the caption for details.

\begin{table}[]
    \captionsetup{font=scriptsize}
    \scriptsize
    \small
    \begin{tabular}{l|lll} % the number of columns needed
        \toprule
        
 & PSNR $\uparrow$ & SSIM $\uparrow$ & LPIPS $\downarrow$ \\
        \midrule
        
 Masked NeRF & 5.91 & 0.70 & 0.40 \\ 
 LaMask & 21.13 & 0.93 & 0.23 \\ 
 SD Text Cond & 13.27 & 0.70 & 0.40 \\ 
 SD Image Cond & 13.71 & 0.75 & 0.31 \\ 
 Extended Attention & 15.20 & 0.78 & 0.30 \\ 
 Grid Prior & 14.90 & 0.82 & 0.27 \\ 
 Joint Multi-View Inpainting & 16.59 & 0.85 & 0.24 \\ 

        \bottomrule
    \end{tabular}
    \caption{\textbf{2D inpainting consistency for data ``chair".} }
    \label{tab:chair}
\end{table}

\begin{table}[]
    \captionsetup{font=scriptsize}
    \scriptsize
    \small
    \begin{tabular}{l|lll} % the number of columns needed
        \toprule
        
 & PSNR $\uparrow$ & SSIM $\uparrow$ & LPIPS $\downarrow$ \\
        \midrule
        
 Masked NeRF & 5.77 & 0.66 & 0.49 \\ 
 LaMask & 16.79 & 0.89 & 0.39 \\ 
 SD Text Cond & 11.17 & 0.72 & 0.29 \\ 
 SD Image Cond & 12.30 & 0.76 & 0.27 \\ 
 Extended Attention & 13.23 & 0.76 & 0.26 \\ 
 Grid Prior & 12.64 & 0.78 & 0.27 \\ 
 Joint Multi-View Inpainting & 13.25 & 0.79 & 0.27 \\ 

        \bottomrule
    \end{tabular}
    \caption{\textbf{2D inpainting consistency for data ``drums".} }
    \label{tab:drums}
\end{table}

\begin{table}[]
    \captionsetup{font=scriptsize}
    \scriptsize
    \small
    \begin{tabular}{l|lll} % the number of columns needed
        \toprule
        
 & PSNR $\uparrow$ & SSIM $\uparrow$ & LPIPS $\downarrow$ \\
        \midrule
        
 Masked NeRF & 6.86 & 0.73 & 0.37 \\ 
 LaMask & 18.50 & 0.91 & 0.22 \\ 
 SD Text Cond & 13.16 & 0.74 & 0.31 \\ 
 SD Image Cond & 13.40 & 0.76 & 0.31 \\ 
 Extended Attention & 14.86 & 0.77 & 0.23 \\ 
 Grid Prior & 14.35 & 0.81 & 0.28 \\ 
 Joint Multi-View Inpainting & 17.24 & 0.85 & 0.21 \\ 

        \bottomrule
    \end{tabular}
    \caption{\textbf{2D inpainting consistency for data ``ficus".} }
    \label{tab:ficus}
\end{table}

\begin{table}[]
    \captionsetup{font=scriptsize}
    \scriptsize
    \small
    \begin{tabular}{l|lll} % the number of columns needed
        \toprule
        
 & PSNR $\uparrow$ & SSIM $\uparrow$ & LPIPS $\downarrow$ \\
        \midrule
        
 Masked NeRF & 8.84 & 0.73 & 0.36 \\ 
 LaMask & 21.29 & 0.92 & 0.15 \\ 
 SD Text Cond & 12.93 & 0.75 & 0.35 \\ 
 SD Image Cond & 15.12 & 0.78 & 0.31 \\ 
 Extended Attention & 15.06 & 0.78 & 0.30 \\ 
 Grid Prior & 15.62 & 0.84 & 0.24 \\ 
 Joint Multi-View Inpainting & 16.69 & 0.86 & 0.24 \\ 

        \bottomrule
    \end{tabular}
    \caption{\textbf{2D inpainting consistency for data ``hotdog".} }
    \label{tab:hotdog}
\end{table}

\begin{table}[]
    \captionsetup{font=scriptsize}
    \scriptsize
    \small
    \begin{tabular}{l|lll} % the number of columns needed
        \toprule
        
 & PSNR $\uparrow$ & SSIM $\uparrow$ & LPIPS $\downarrow$ \\
        \midrule
        
 Masked NeRF & 8.80 & 0.73 & 0.33 \\ 
 LaMask & 17.84 & 0.84 & 0.18 \\ 
 SD Text Cond & 13.01 & 0.75 & 0.27 \\ 
 SD Image Cond & 14.79 & 0.79 & 0.22 \\ 
 Extended Attention & 15.42 & 0.79 & 0.22 \\ 
 Grid Prior & 14.84 & 0.81 & 0.21 \\ 
 Joint Multi-View Inpainting & 17.89 & 0.84 & 0.18 \\ 

        \bottomrule
    \end{tabular}
    \caption{\textbf{2D inpainting consistency for data ``lego".} }
    \label{tab:lego}
\end{table}

\begin{table}[]
    \captionsetup{font=scriptsize}
    \scriptsize
    \small
    \begin{tabular}{l|lll} % the number of columns needed
        \toprule
        
 & PSNR $\uparrow$ & SSIM $\uparrow$ & LPIPS $\downarrow$ \\
        \midrule
        
 Masked NeRF & 8.28 & 0.72 & 0.33 \\ 
 LaMask & 18.23 & 0.88 & 0.16 \\ 
 SD Text Cond & 12.80 & 0.74 & 0.29 \\ 
 SD Image Cond & 14.53 & 0.78 & 0.25 \\ 
 Extended Attention & 12.85 & 0.77 & 0.29 \\ 
 Grid Prior & 14.25 & 0.80 & 0.22 \\ 
 Joint Multi-View Inpainting & 14.53 & 0.80 & 0.23 \\ 

        \bottomrule
    \end{tabular}
    \caption{\textbf{2D inpainting consistency for data ``materials".} }
    \label{tab:materials}
\end{table}

\begin{table}[]
    \captionsetup{font=scriptsize}
    \scriptsize
    \small
    \begin{tabular}{l|lll} % the number of columns needed
        \toprule
        
 & PSNR $\uparrow$ & SSIM $\uparrow$ & LPIPS $\downarrow$ \\
        \midrule
        
 Masked NeRF & 6.94 & 0.73 & 0.35 \\ 
 LaMask & 21.11 & 0.92 & 0.13 \\ 
 SD Text Cond & 10.97 & 0.72 & 0.31 \\ 
 SD Image Cond & 13.40 & 0.77 & 0.28 \\ 
 Extended Attention & 14.26 & 0.78 & 0.27 \\ 
 Grid Prior & 12.41 & 0.80 & 0.26 \\ 
 Joint Multi-View Inpainting & 13.26 & 0.81 & 0.27 \\ 

        \bottomrule
    \end{tabular}
    \caption{\textbf{2D inpainting consistency for data ``mic".} }
    \label{tab:mic}
\end{table}

\begin{table}[]
    \captionsetup{font=scriptsize}
    \scriptsize
    \small
    \begin{tabular}{l|lll} % the number of columns needed
        \toprule
        
 & PSNR $\uparrow$ & SSIM $\uparrow$ & LPIPS $\downarrow$ \\
        \midrule
        
 Masked NeRF & 10.66 & 0.69 & 0.35 \\ 
 LaMask & 21.76 & 0.82 & 0.18 \\ 
 SD Text Cond & 13.14 & 0.72 & 0.31 \\ 
 SD Image Cond & 15.95 & 0.74 & 0.29 \\ 
 Extended Attention & 15.69 & 0.74 & 0.29 \\ 
 Grid Prior & 16.41 & 0.78 & 0.27 \\ 
 Joint Multi-View Inpainting & 17.64 & 0.80 & 0.24 \\ 

        \bottomrule
    \end{tabular}
    \caption{\textbf{2D inpainting consistency for data ``ship".} }
    \label{tab:ship}
\end{table}
\begin{table}[]
    \captionsetup{font=scriptsize}
    \scriptsize
    \footnotesize
    \begin{tabular}{l|lll|ll} % the number of columns needed
        \toprule
        
 & PSNR $\uparrow$ & SSIM $\uparrow$ & LPIPS $\downarrow$ & MUSIQ $\uparrow$ & Corrs $\uparrow$ \\
        \midrule
        
 Masked NeRF & 12.45 & 0.81 & 0.30 & 3.82 & 330 \\ 
 LaMask & 27.43 & 0.95 & 0.03 & 3.72 & 179 \\ 
 SD Image Cond & 17.42 & 0.88 & 0.15 & 3.55 & 305 \\ 
 Inpaint + DU & 25.42 & 0.95 & 0.06 & 3.96 & 260 \\ 
 Ours w/o depth & 29.80 & 0.96 & 0.03 & 3.91 & 218 \\ 
 Ours & 29.71 & 0.96 & 0.03 & 3.92 & 213 \\ 

        \bottomrule
    \end{tabular}
    \caption{\textbf{Quantitative NeRF baselines for data ``cat".} }
    \label{tab:jerrypillow}
\end{table}

\begin{table}[]
    \captionsetup{font=scriptsize}
    \scriptsize
    \footnotesize
    \begin{tabular}{l|lll|ll} % the number of columns needed
        \toprule
        
 & PSNR $\uparrow$ & SSIM $\uparrow$ & LPIPS $\downarrow$ & MUSIQ $\uparrow$ & Corrs $\uparrow$ \\
        \midrule
        
 Masked NeRF & 19.16 & 0.90 & 0.15 & 3.65 & 164 \\ 
 LaMask & 31.66 & 0.96 & 0.03 & 3.66 & 140 \\ 
 SD Image Cond & 21.96 & 0.89 & 0.12 & 3.62 & 182 \\ 
 Inpaint + DU & 28.40 & 0.94 & 0.07 & 3.78 & 151 \\ 
 Ours w/o depth & 29.76 & 0.94 & 0.06 & 3.65 & 154 \\ 
 Ours & 29.74 & 0.93 & 0.07 & 3.69 & 146 \\ 

        \bottomrule
    \end{tabular}
    \caption{\textbf{Quantitative NeRF baselines for data ``turtle".} }
    \label{tab:turtle}
\end{table}

\begin{table}[]
    \captionsetup{font=scriptsize}
    \scriptsize
    \footnotesize
    \begin{tabular}{l|lll|ll} % the number of columns needed
        \toprule
        
 & PSNR $\uparrow$ & SSIM $\uparrow$ & LPIPS $\downarrow$ & MUSIQ $\uparrow$ & Corrs $\uparrow$ \\
        \midrule
        
 Masked NeRF & 14.20 & 0.70 & 0.34 & 3.95 & 1083 \\ 
 LaMask & 25.45 & 0.76 & 0.09 & 4.03 & 1040 \\ 
 SD Image Cond & 24.23 & 0.77 & 0.12 & 4.00 & 1024 \\ 
 Inpaint + DU & 25.15 & 0.76 & 0.13 & 3.97 & 1041 \\ 
 Ours w/o depth & 26.68 & 0.82 & 0.12 & 4.01 & 1019 \\ 
 Ours & 26.49 & 0.81 & 0.13 & 4.02 & 1038 \\ 

        \bottomrule
    \end{tabular}
    \caption{\textbf{Quantitative NeRF baselines for data ``drawing".} }
    \label{tab:drawing}
\end{table}

\begin{table}[]
    \captionsetup{font=scriptsize}
    \scriptsize
    \footnotesize
    \begin{tabular}{l|lll|ll} % the number of columns needed
        \toprule
        
 & PSNR $\uparrow$ & SSIM $\uparrow$ & LPIPS $\downarrow$ & MUSIQ $\uparrow$ & Corrs $\uparrow$ \\
        \midrule
        
 Masked NeRF & 11.68 & 0.79 & 0.25 & 3.81 & 214 \\ 
 LaMask & 24.50 & 0.95 & 0.05 & 4.02 & 195 \\ 
 SD Image Cond & 16.45 & 0.89 & 0.12 & 3.66 & 196 \\ 
 Inpaint + DU & 20.42 & 0.90 & 0.10 & 3.80 & 228 \\ 
 Ours w/o depth & 28.76 & 0.96 & 0.03 & 3.84 & 211 \\ 
 Ours & 29.37 & 0.96 & 0.03 & 3.86 & 176 \\ 

        \bottomrule
    \end{tabular}
    \caption{\textbf{Quantitative NeRF baselines for data ``boot".} }
    \label{tab:boot}
\end{table}

\begin{table}[]
    \captionsetup{font=scriptsize}
    \scriptsize
    \footnotesize
    \begin{tabular}{l|lll|ll} % the number of columns needed
        \toprule
        
 & PSNR $\uparrow$ & SSIM $\uparrow$ & LPIPS $\downarrow$ & MUSIQ $\uparrow$ & Corrs $\uparrow$ \\
        \midrule
        
 Masked NeRF & 16.86 & 0.93 & 0.10 & 3.82 & 245 \\ 
 LaMask & 27.41 & 0.96 & 0.02 & 3.71 & 200 \\ 
 SD Image Cond & 24.58 & 0.95 & 0.04 & 3.70 & 261 \\ 
 Inpaint + DU & 27.92 & 0.95 & 0.03 & 3.70 & 245 \\ 
 Ours w/o depth & 28.20 & 0.96 & 0.03 & 3.72 & 259 \\ 
 Ours & 28.13 & 0.96 & 0.03 & 3.72 & 264 \\ 

        \bottomrule
    \end{tabular}
    \caption{\textbf{Quantitative NeRF baselines for data ``bear".} }
    \label{tab:bear}
\end{table}

\begin{table}[]
    \captionsetup{font=scriptsize}
    \scriptsize
    \footnotesize
    \begin{tabular}{l|lll|ll} % the number of columns needed
        \toprule
        
 & PSNR $\uparrow$ & SSIM $\uparrow$ & LPIPS $\downarrow$ & MUSIQ $\uparrow$ & Corrs $\uparrow$ \\
        \midrule
        
 Masked NeRF & 11.96 & 0.75 & 0.29 & 3.80 & 193 \\ 
 LaMask & 27.73 & 0.97 & 0.04 & 3.84 & 148 \\ 
 SD Image Cond & 19.07 & 0.87 & 0.15 & 3.54 & 235 \\ 
 Inpaint + DU & 28.76 & 0.95 & 0.05 & 3.69 & 210 \\ 
 Ours w/o depth & 27.75 & 0.95 & 0.04 & 3.63 & 245 \\ 
 Ours & 27.48 & 0.95 & 0.05 & 3.62 & 232 \\ 

        \bottomrule
    \end{tabular}
    \caption{\textbf{Quantitative NeRF baselines for data ``dumptruck".} }
    \label{tab:dumptruck}
\end{table}

\begin{table}[]
    \captionsetup{font=scriptsize}
    \scriptsize
    \footnotesize
    \begin{tabular}{l|lll|ll} % the number of columns needed
        \toprule
        
 & PSNR $\uparrow$ & SSIM $\uparrow$ & LPIPS $\downarrow$ & MUSIQ $\uparrow$ & Corrs $\uparrow$ \\
        \midrule
        
 Masked NeRF & 16.23 & 0.70 & 0.32 & 3.65 & 934 \\ 
 LaMask & 27.25 & 0.86 & 0.09 & 3.84 & 865 \\ 
 SD Image Cond & 21.58 & 0.81 & 0.14 & 3.98 & 852 \\ 
 Inpaint + DU & 27.75 & 0.87 & 0.10 & 3.83 & 812 \\ 
 Ours w/o depth & 29.54 & 0.90 & 0.07 & 3.65 & 980 \\ 
 Ours & 29.11 & 0.88 & 0.07 & 3.69 & 1059 \\ 

        \bottomrule
    \end{tabular}
    \caption{\textbf{Quantitative NeRF baselines for data ``norway".} }
    \label{tab:norway}
\end{table}

\begin{table}[]
    \captionsetup{font=scriptsize}
    \scriptsize
    \footnotesize
    \begin{tabular}{l|lll|ll} % the number of columns needed
        \toprule
        
 & PSNR $\uparrow$ & SSIM $\uparrow$ & LPIPS $\downarrow$ & MUSIQ $\uparrow$ & Corrs $\uparrow$ \\
        \midrule
        
 Masked NeRF & 16.41 & 0.80 & 0.19 & 3.46 & 1766 \\ 
 LaMask & 24.82 & 0.84 & 0.04 & 3.41 & 1833 \\ 
 SD Image Cond & 22.48 & 0.83 & 0.07 & 3.45 & 1769 \\ 
 Inpaint + DU & 24.04 & 0.84 & 0.06 & 3.47 & 1805 \\ 
 Ours w/o depth & 24.38 & 0.85 & 0.04 & 3.39 & 1847 \\ 
 Ours & 23.93 & 0.83 & 0.05 & 3.40 & 1924 \\ 

        \bottomrule
    \end{tabular}
    \caption{\textbf{Quantitative NeRF baselines for data ``backpack".} }
    \label{tab:backpack}
\end{table}

\begin{table}[]
    \captionsetup{font=scriptsize}
    \scriptsize
    \footnotesize
    \begin{tabular}{l|lll|ll} % the number of columns needed
        \toprule
        
 & PSNR $\uparrow$ & SSIM $\uparrow$ & LPIPS $\downarrow$ & MUSIQ $\uparrow$ & Corrs $\uparrow$ \\
        \midrule
        
 Masked NeRF & 13.37 & 0.69 & 0.37 & 3.50 & 981 \\ 
 LaMask & 25.63 & 0.84 & 0.11 & 3.59 & 1053 \\ 
 SD Image Cond & 24.31 & 0.83 & 0.12 & 3.66 & 1023 \\ 
 Inpaint + DU & 27.13 & 0.83 & 0.13 & 3.69 & 1078 \\ 
 Ours w/o depth & 28.87 & 0.88 & 0.09 & 3.66 & 1120 \\ 
 Ours & 28.78 & 0.88 & 0.09 & 3.66 & 1137 \\ 

        \bottomrule
    \end{tabular}
    \caption{\textbf{Quantitative NeRF baselines for data ``billiards".} }
    \label{tab:billiards}
\end{table}

\begin{table}[]
    \captionsetup{font=scriptsize}
    \scriptsize
    \footnotesize
    \begin{tabular}{l|lll|ll} % the number of columns needed
        \toprule
        
 & PSNR $\uparrow$ & SSIM $\uparrow$ & LPIPS $\downarrow$ & MUSIQ $\uparrow$ & Corrs $\uparrow$ \\
        \midrule
        
 Masked NeRF & 14.74 & 0.75 & 0.31 & 3.65 & 839 \\ 
 LaMask & 32.02 & 0.93 & 0.03 & 3.80 & 779 \\ 
 SD Image Cond & 28.26 & 0.90 & 0.07 & 3.64 & 799 \\ 
 Inpaint + DU & 31.05 & 0.92 & 0.05 & 3.70 & 766 \\ 
 Ours w/o depth & 30.35 & 0.93 & 0.05 & 3.68 & 764 \\ 
 Ours & 30.08 & 0.93 & 0.05 & 3.69 & 768 \\ 

        \bottomrule
    \end{tabular}
    \caption{\textbf{Quantitative NeRF baselines for data ``office".} }
    \label{tab:office}
\end{table}

\end{document}